\pdfoutput=1

\documentclass[11pt]{article}

\usepackage[final]{acl}

\usepackage{times}
\usepackage{latexsym}

\usepackage[T1]{fontenc}

\usepackage[utf8]{inputenc}

\usepackage{microtype}

\usepackage{inconsolata}

\usepackage{graphicx}
\usepackage{multirow}

\newcommand{\bhline}{\noalign{\hrule height 1.2pt}}

\title{Do LLMs and Humans Find the Same Questions Difficult? \\
A Case Study on Japanese Quiz Answering}

\author{
    \textbf{Naoya Sugiura \hspace{4ex} Kosuke Yamada \hspace{4ex} Yasuhiro Ogawa}\\
    \textbf{Katsuhiko Toyama \hspace{4ex} Ryohei Sasano}\\
    Graduate School of Informatics, Nagoya University
}

\begin{document}
\maketitle
\begin{abstract}
LLMs have achieved performance that surpasses humans in many NLP tasks.
However, it remains unclear whether problems that are difficult for humans are also difficult for LLMs.
This study investigates how the difficulty of quizzes in a buzzer setting differs between LLMs and humans.
Specifically, we first collect Japanese quiz data including questions, answers, and correct response rate of humans, 
then prompted LLMs to answer the quizzes under several settings, and compare their correct answer rate to that of humans from two analytical perspectives.
The experimental results showed that, compared to humans, LLMs struggle more with quizzes whose correct answers are not covered by Wikipedia entries, and also have difficulty with questions that require numerical answers.
\end{abstract}

\section{Introduction}
Large language models (LLMs) have demonstrated performance superior to that of humans in a variety of NLP tasks~\cite{openai2024gpt4technicalreport, touvron2023llama2openfoundation, jiang2023mistral7b}.
This is also true for quiz answering, an open-domain question answering task that requires general knowledge and understanding of current events~\cite{ariyamatomoki2024}.
For example, in the Japanese quiz competition AIO,\footnote{\url{https://sites.google.com/view/project-aio}} LLM-based systems achieved a correct response (CR) rate of over 90\%, which significantly exceeds human performance.

Interestingly, despite the high performance of LLMs, their error patterns have been observed to differ from those of humans.
Even with questions considered easy for most humans, LLMs occasionally fail to produce correct answers.
Table \ref{tab:wrong_example} presents an example.\footnote{The actual example is in Japanese, but its English translation is provided for legibility.} 
While the human CR rate for this question reaches 67\%, GPT-4o often responds incorrectly.
It suggests that the concept of \textsl{difficulty} for humans does not necessarily align with that for LLMs, which can be a critical issue for services that rely on LLMs because most people using LLM expect them to be superior to humans. 
Understanding these differences can provide valuable insights into the characteristics and limitations of LLMs. 

Existing studies on the analysis of quiz answering by LLMs have focused on cognitive aspects of quiz answering, including the impact of question format on LLM reasoning \cite{wallace-boyd-graber-2018-trick, wallace-etal-2019-trick,NLP2024_Parallel} and human cognitive processes \cite{hayoshi_hukakujitsu}.
In contrast, this study focuses on comparing and analyzing the error patterns of LLMs and humans. 
Specifically, we collect Japanese buzzer quiz data accompanied by human CR rates and categorize the data from two perspectives: whether the answer has a corresponding Wikipedia entry and what types of characters\footnote{For the types of characters in Japanese, see Appendix \ref{app:Japanese_charactesrs}.} appear in the answer.
We then analyze how CR rates differ across categories to clarify the differences in quiz difficulty between LLMs and humans.

\begin{table}[t!]
\centering
\small
\begin{tabular}{@{\ }l@{\ }} \bhline
  \rule{0pt}{2.0ex}
 \textbf{Question:} What is the portable lighting device, whose \\ 
 name is derived from the Ancient Greek word for 'torch,'  \\ 
 and is commonly used for camping and other activities? \\ 
\textbf{Correct Response Rate of Humans}: 67.0\% \\
 \textbf{Answer:} Lantern\ \ \textbf{Model Output:} {\color{red}Torch}\ \ \ \ \\ \bhline
\end{tabular}
\caption{An example quiz that is easy for humans but frequently answered incorrectly by GPT-4o.}
\label{tab:wrong_example}
\end{table}

\section{Quiz Answering and Dataset}
We introduce previous studies on quiz answering and the quiz data collected in this study.

\subsection{Quiz Answering}
Quiz answering has been studied as one of the open-domain question answering (ODQA) tasks and is characterized by its short, single-question format \cite{NLP2024_meisu}.
Various methods used in ODQA have been applied to quiz answering, including generative approaches \cite{brown2020languagemodelsfewshotlearners, izacard-grave-2021-leveraging, 10.1162/tacl_a_00415} and hybrid approaches that combine text retrieval with text generation \cite{karpukhin-etal-2020-dense, yamada-etal-2021-efficient, Raffel2019ExploringTL}.
Recent studies have conducted cognitive analyses of LLM performance on quiz answering \cite{NLP2024_Parallel} and have explored characteristics which make questions difficult for LLMs \cite{wallace-etal-2019-trick, wallace-boyd-graber-2018-trick}

Existing study has also explored buzzer quizzes, in which participants are allowed to answer before the question is fully presented.
Answering such quizzes is more challenging for LLMs, as they must generate responses before receiving the entire question.
Several researchers have worked on building answering systems for incomplete questions \cite{boyd-graber-etal-2012-besting, iyyer-etal-2014-neural, sugiura-etal-2023-building} and some have analyzed human decision-making and strategy shifts when dealing with incomplete information \cite{hayoshi_hukakujitsu}. 
These studies focus on building automated buzzer-quiz answering systems and analyzing the human answering process, rather than comparing LLMs and humans.

\subsection{Quiz Dataset with human CR rates}
\label{subsec:dataset_with_human_CR rate}
We collected quiz data and their corresponding human CR rates from the quiz game application Minna-de Hayaoshi Quiz,\footnote{\url{https://minhaya.com/}} using data from external resources.\footnote{\url{https://mininome.com/}}$^,$\footnote{\url{https://raityo.com/}}
In this application, players create quiz questions and answers, which are then reviewed and approved by the administrators, resulting in a high-quality and diverse set of quizzes.
This application provides a human CR rate for each question, defined as the proportion of correct respondents to the total number of respondents.
The game employs a buzzer-quiz setting, allowing players to answer before the entire question is presented.

\section{Analysis Methods}
In this section, we describe the methods for quiz answering using LLMs, the correctness evaluation process, and the perspectives used for analysis and comparison.

\subsection{Quiz Answering by LLMs}
To determine the base LLM, we first constructed and compared quiz answering systems using three types of LLMs under the setting where the entire question was provided.
Specifically, we used GPT-4o,\footnote{\href{https://platform.openai.com/docs/models\#gpt-4o}{gpt-4o-2024-08-06}} a model primarily pre-trained in English; Swallow~70B,\footnote{\href{https://huggingface.co/tokyotech-llm/Swallow-70b-hf}{tokyotech-llm/Swallow-70b-hf}} a model that has undergone continual pre-training from the Llama 2 with 90B Japanese tokens; and Sarashina2~70B,\footnote{\href{https://huggingface.co/sbintuitions/sarashina2-70b}{sbintuitions/sarashina2-70b}} a model primarily pre-trained in Japanese, using 1T Japanese tokens. 
We applied two types of additional training to each model: few-shot learning and fine-tuning using QLoRA \cite{dettmers2023qloraefficientfinetuningquantized}.
For each language model, we adopted the setting that yielded the highest CR rate.
Specifically, we used the model trained with 5-shot learning for GPT-4o, the fine-tuned model for Swallow, and the model without any additional training for Sarashina2.

In order to simulate the behavior of LLM-based models in a buzzer-quiz setting, we designed experiments where the model received only a part of the question text.
Specifically, we used GPT-4o, which showed the best performance when the entire question was provided, as the base LLM, and conducted experiments where the model was given only the first 75\%, 50\%, or 25\% of each question.

\subsection{Correctness Evaluation Process}
\label{subsec:evalutation}
The correctness of model responses was determined using the following two-step process:

\begin{description}
\item[Step 1:] Six types of marks, i.e., `(', `)',  `[', `]', `.', and `=', are removed from both the model response and the answer of the quiz. 
If the resulting strings match exactly, the response is considered correct.

\item[Step 2:] If the model response is not classified as correct in Step 1, 
it is evaluated by a spelling variation normalizer based on GPT-4o (see Appendix \ref{app:re-evaluationer}). 
If the module determines that the response is a valid spelling variation of the answer, it is considered correct.
\end{description}

To confirm the effect of the normalizer, we also calculated the CR rate using only Step 1.

\subsection{Perspectives for Analysis}
\label{subsec:point_of_analysis}
We categorize quizzes from two perspectives: the existence of a Wikipedia entry for the answer and the character types used in the answer.
We compare the error patterns between LLMs and humans based on the categorized results.

\paragraph{Wikipedia Entry}
LLMs are considered to possess extensive encyclopedic knowledge, particularly the kind of information typically found in Wikipedia entries.
In addition, LLMs are trained on a vast amount of web data, with Wikipedia being one of the most prominent sources.
Therefore, quizzes whose answers correspond to a Wikipedia entry are likely to be easier for LLMs to answer.
In contrast, since humans do not possess exhaustive encyclopedic knowledge, whether the answer is present as a Wikipedia entry is unlikely to significantly affect human CR rates.
To test this hypothesis, we conduct an analysis based on whether the quiz answers are included in Wikipedia entries.

We determine the existence of a Wikipedia entry by extracting all article titles from a dump of the Wikipedia dated October 1, 2024.
Before making comparisons, we preprocessed the titles by removing the six types of punctuation marks described in Section \ref{subsec:evalutation}.
Wikipedia entries can appear in two forms: some have a dedicated article, while others exist only as redirects to another article.
We distinguish between these two cases in our analysis.

\paragraph{Character Type}
The Japanese writing system often reflects the origin of a word through the script in which it is written.
Specifically, loanwords tend to appear in katakana, acronyms in the Latin script are typically retained in their original form, and native Japanese words are usually expressed in either hiragana or kanji.
Additionally, some quiz answers are numerical, which often indicates that the question involves arithmetic, enumeration, or other forms of numerical reasoning, potentially making them more challenging for LLMs.

Given that the character type of the answer may influence the CR rates of LLMs, this study categorizes quizzes based on the characters used in the answer.
Using regular expressions, we determine whether the answer falls into one of four categories: numerals, Latin letters, katakana, or hiragana/kanji. 
Questions are then analyzed based on these results.

\section{Results and Discussions}
We present the CR rate of each model and analyze the results based on the perspectives. 
We then discuss these findings in detail.

In Tables \ref{tab:result_wikipedia} and \ref{tab:result_character}, which show the results for each perspective, the values in parentheses in the header row indicate the number of quizzes classified into each corresponding category.
The values in parentheses elsewhere in the table cells represent the CR rate (\%) before applying normalization for spelling variations. 
The row labeled ``Human'' indicates the average human CR rates (\%). 

In addition, for the results of GPT-4o under the buzzer-quiz setting, where only the first 75\%/50\%/25\% of the question text is provided, the model is denoted as ``GPT-4o (75\%/50\%/25\%).''

\begin{table}[t!]
\centering
\footnotesize
\begin{tabular}{@{ }l@{\ }|@{\ }c@{\ \ }c@{\ \ }c@{\ }|@{\ }c@{}} \bhline
\multirow{2}{*}{Model} & Article & Redirect & None & Overall\\ 
& (3,042) & (486) & (440) & (3,968) \\ \hline
Swallow~70B & 91.7 \scriptsize{(84.6)} & 72.6 \scriptsize{(42.6)} & 70.7 \scriptsize{(33.1)} & 87.0 \scriptsize{(73.7)} \\ 
Sarashina2~70B & 93.8 \scriptsize{(84.7)} & 88.9 \scriptsize{(62.8)} & 81.6 \scriptsize{(48.8)} & 91.8 \scriptsize{(78.0)} \\ 
GPT-4o & \textbf{96.6 \scriptsize{(88.7)}} & \textbf{92.2 \scriptsize{(71.8)}} & \textbf{87.3 \scriptsize{(61.5)}} & \textbf{95.0 \scriptsize{(83.6)}} \\ 
GPT-4o (75\%) & 93.9 \scriptsize{(85.2)} & 89.3 \scriptsize{(66.9)} & 82.3 \scriptsize{(54.2)} & 92.0 \scriptsize{(83.6)} \\
GPT-4o (50\%) & 78.1 \scriptsize{(69.1)} & 75.5 \scriptsize{(53.9)} & 61.7 \scriptsize{(38.8)} & 76.0 \scriptsize{(63.8)} \\ 
GPT-4o (25\%) & 47.0 \scriptsize{(41.6)} & 42.6 \scriptsize{(27.2)} & 26.3 \scriptsize{(13.2)} & 44.2 \scriptsize{(36.6)} \\ 
\hline
Human & 41.9 & 41.1 & 40.5 & 41.6 \\ \bhline
\end{tabular}
\caption{CR rates for Wikipedia entries of answers.
``Article'', ``Redirect'', and  ``None'' refer to cases where the answer has a corresponding Wikipedia entry, where the answer exists as a redirect to another entry, and where the answer does not have an entry, respectively.}
\label{tab:result_wikipedia}
\end{table}

\subsection{Overall Trend of CR Rate}
The rightmost column of Table \ref{tab:result_wikipedia} presents the overall CR rate of each model, as well as the average CR rate of human participants.
Among the models, GPT-4o achieved the highest CR rate.
Despite the buzzer-quiz setting, GPT-4o (75\%) maintained a high CR rate and even when provided with only 25\% of the question text, GPT-4o slightly outperformed the average human CR rate.
A detailed analysis of the relationship between human and LLM CR rates is provided in Appendix \ref{app:comparizon_accuracy}.

\subsection{Effect of Wikipedia Entry of Answer}
Table \ref{tab:result_wikipedia} presents the CR rates for each category based on the presence or absence of a Wikipedia entry.
Across all models, the CR rate consistently decreased in the following order: Article, Redirect, None.
In contrast, the difference in human CR rate across these categories was approximately one percentage point, which is smaller than the CR rate gap observed in the models. 
This finding supports the hypothesis presented in Section \ref{subsec:point_of_analysis}, indicating that Wikipedia serves as an essential knowledge resource during pre-training of LLMs. 
Another possible explanation for the higher CR rate in questions where the answer has a Wikipedia entry is that such words tend to be more commonly used. 
This could result in a higher frequency of occurrence in the training corpus, leading to an increased CR rate.

Next, when comparing the improvement in CR rate before and after applying the normalization module, we can observe that questions without Wikipedia entries show relatively large improvements compared to questions with entries.
This indicates a much higher prevalence of spelling variations in these cases. 
The lower frequency of spelling variations in questions with Wikipedia entries can likely be attributed to the fact that words with Wikipedia entries tend to have standardized spellings. 
Additionally, these standardized forms are often registered in widely used conversion systems and predictive text dictionaries, further reducing spelling variation in their written forms.

When only the initial portion of each quiz question was provided as input to the GPT-4o model, the relative performance ranking across categories remained unchanged compared to when the entire text was used. However, the degree of performance degradation tended to be greater in categories that originally exhibited lower accuracy.

\begin{table}[t!]
\centering
\small
\begin{tabular}{@{}l@{\ \ }|@{\ \ }c@{\ \ }c@{\ \ }c@{\ \ }c@{}} \bhline
\multirow{3}{*}{Model} & \multirow{2}{*}{Numeral} & Latin & \multirow{2}{*}{Katakana} & Hiragana \\[-2pt] 
& & letters & & Kanji \\ 
& (141) & (153) & (1,835) & (1,838) \\ \hline
Swallow~70B & 78.0 \scriptsize{(43.3)} & 69.3 \scriptsize{(51.0)} & 90.8 \scriptsize{(78.9)} & 85.5 \scriptsize{(73.0)} \\
Sarashina2~70B & 83.7 \scriptsize{(49.6)} & 93.5 \scriptsize{(69.9)} & 93.7 \scriptsize{(80.6)} & 90.6 \scriptsize{(78.5)} \\
GPT-4o & \textbf{92.2 \scriptsize{(67.4)}} & \textbf{98.7 \scriptsize{(81.0)}} & \textbf{96.6 \scriptsize{(84.7)}} & \textbf{93.4 \scriptsize{(84.1)}} \\ 
GPT-4o (75\%) & 92.2 \scriptsize{(65.2)} & 95.4 \scriptsize{(75.2)} & 94.6 \scriptsize{(81.5)} & 89.2 \scriptsize{(79.1)} \\ 
GPT-4o (50\%) & 66.7 \scriptsize{(41.8)} & 83.0 \scriptsize{(65.4)} & 81.0 \scriptsize{(68.1)} & 71.2 \scriptsize{(61.2)} \\
GPT-4o (25\%) & 18.4 \scriptsize{(11.3)} & 52.9 \scriptsize{(40.5)} & 49.8 \scriptsize{(40.9)} & 39.9 \scriptsize{(34.1)} \\
\hline
Human & 39.7 & 45.6 & 40.1 & 42.9 \\ \bhline
\end{tabular}
\caption{CR rates for each character type of the answer.}
\label{tab:result_character}
\end{table}

\subsection{Effect of Character Type of Answer}
Table \ref{tab:result_character} presents the CR rates for each category based on the character type of the answer.
Across all models and conditions, questions with answers written in katakana had relatively high CR rates and questions with answers written in numerals had relatively low CR rates.
Compared to the overall human CR rate of 41.6 reported in Table 2, the human CR rates for these two categories are slightly lower, but the differences are limited.
This indicates that LLMs find questions that require numerical answers more difficult than humans do.
We further analyze trends specific to numerical answers.
Many of these questions involve enumeration, such as ``How many of the 50 U.S. states have a coastline?''
Solving such questions requires knowledge of all 50 states of U.S and whether they have a coastline. 
The requirement for integrating multiple types of knowledge may increase the difficulty of these questions for LLMs.

The trend in the CR rate for the questions with answers written in Latin characters, for which the humans had the highest CR rate, differed depending on the model, resulting in the highest CR rate for GPT-4o, and the lowest CR rate for Swallow 70B.
Answers in katakana and Latin letters are mostly loanwords, whereas answers in hiragana and kanji are typically native Japanese words. 
Thus, we predicted that the CR rate for questions with answers written in Latin letters or katakana would be low for the models that used Japanese in their training, i.e., Swallow and Sarashina2. 
However, even for these models, while the CR rate for questions with answers in Latin letters was relatively low, the CR rate for questions with answers in katakana was still high.

The trend observed when the input to GPT-4o was reduced was consistent with the analysis based on Wikipedia entries.
That is, the relative performance ranking across categories remained unchanged compared to when the entire text was used, but the degree of performance degradation tended to be greater in categories that originally exhibited lower accuracy.

\section{Conclusion}
In this study, we analyzed and compared the differences in error patterns between LLMs and humans in Japanese quizzes, investigating how the perception of \textsl{difficulty} differs between the two.
Our analysis of Wikipedia entry-related factors revealed that the CR rate of LLMs dropped greatly for questions where the answer lacked a corresponding Wikipedia entry, whereas the human CR rate remained largely unaffected. 
This result indicates that Wikipedia serves as a major knowledge resource for LLMs.
Regarding the character type, we found that the CR rate was relatively low when the answer was written in numeral.
We confirmed that the CR rates for questions with answers written in Latin letters, for which the humans had the highest CR rate, vary greatly depending on the data used for training.
These findings indicates that certain quiz characteristics affect only LLMs, which likely contribute to the difference in perceived difficulty between LLMs and humans.

\section*{Limitations}
The human CR rate data used in this study was calculated in a buzzer-quiz setting, and there is a possibility that it will differ from the CR rates when the input to the LLMs is uniformly truncated by a fixed proportion, regardless of the content of the quiz questions.
In addition, the skill levels of the players used to calculate the CR rate remain unknown. 
To enable a more precise comparison, it is necessary to conduct controlled human experiments and directly measure the CR rate, which remains a challenge for future work.
All of the experiments in this study were conducted in Japanese, and it is difficult to apply the analysis based on character type to languages other than Japanese, which use multiple character types.


\bibliography{custom}

\appendix

\section{Types of Characters in Japanese}
\label{app:Japanese_charactesrs}
The Japanese writing system combines Chinese characters, known as kanji, with two unique syllabaries: hiragana and katakana. 
Kanji and hiragana are the most widely used, and katakana is mainly used to write loanwords.
Latin letters are also used to represent words such as English acronyms and product names.
The numeral system uses mostly Arabic numerals, but also traditional Chinese numerals.

\section{Spelling Variation Normalizer}
\label{app:re-evaluationer}
The spelling variation normalizer is a binary classifier that takes as input the quiz question, the answer, and the model's response. 
If it determines that the response was unfairly marked incorrect due to a spelling variation, it returns True; if it judges the response to be a genuine incorrect answer, it returns False.
This classifier was built using GPT-4o with few-shot learning. The number of shots was determined by testing with manually created evaluation data, and the best-performing setting was found to be 6-shot.
The confusion matrix for the test results under the 6-shot setting is presented in Table \ref{tab:confusion_matrix}.
Although there are existing studies on answer validation systems \cite{}, the results shown in Table \ref{tab:confusion_matrix} indicate that the system used in this study achieves sufficient performance under our experimental setting.
In addition, the example outputs are shown in Table \ref{tab:yure_example}.

\section{Error Rates of Humans and Models}
\label{app:comparizon_accuracy}

\begin{figure*}[t]
\centering
\includegraphics[width=.95\linewidth]{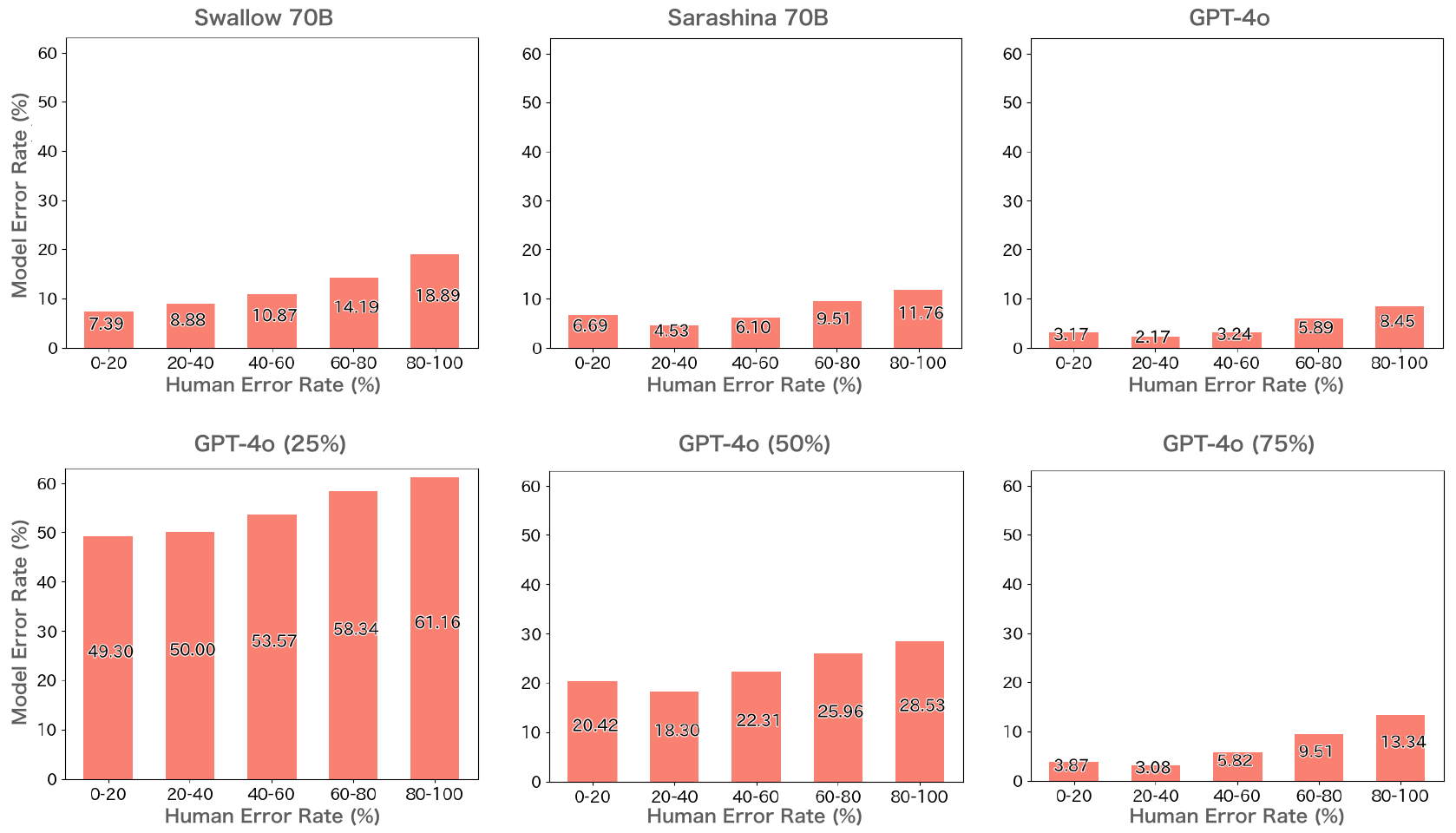}
\caption{Model error rate relative to human error rate.}
\label{fig:err_rate_comparizon}
\end{figure*}

Figure \ref{fig:err_rate_comparizon} presents a graph comparing the model error rate against the human error rate.
As a whole, a positive correlation was observed between human and model error rates. 
Interestingly, however, in the half of the settings, the error rate of the LLM-based models for questions with the human error rate between 20\% and 40\% was the highest.
This indicates that there are problems that are very easy for people, but difficult for LLM-based models.

\begin{table}[t]
\centering
\small
\begin{tabular}{cc|cc}\bhline
&       & \multicolumn{2}{c}{Actual} \\ 
&       & True & False \\ \hline
\rule{0pt}{2.7ex}
\multirow{2}{*}{Predict} &  True  &  \multicolumn{1}{r}{\normalsize 228}  &  
\multicolumn{1}{r}{\normalsize 6}     \\  
\rule{0pt}{3ex}
& False & \multicolumn{1}{r}{\normalsize 1}    & \multicolumn{1}{r}{\normalsize 265}  \\ \bhline
\end{tabular}
\caption{Test results of the spelling variation normalizer in the 6-shot setting.}
\label{tab:confusion_matrix}
\end{table}

\begin{table}[t]
\centering
\footnotesize
\begin{tabular}{@{\ }l@{}} \bhline
\rule{0pt}{2.0ex} 
 \textbf{Question}: Which English physicist is known for \\ discovering the law of universal gravitation after observing  \\an apple fall from a tree?\\
 \textbf{Answer:} Isaac Newton\ \ \textbf{Model Output:} Sir Isaac Newton\ \ \  \\ 
 \textbf{Normalizer Output:} True \\\hline
 \rule{0pt}{2.0ex}
 \textbf{Question:}  How many of the 50 U.S. states have \\ 
 a coastline?\\
 \textbf{Answer:} 24\ \ \textbf{Model Output:} 24 states \ \ \  \\ 
 \textbf{Normalizer Output:} True \\\hline
  \rule{0pt}{2.0ex}
 \textbf{Question:}  Who served as the first Vice President of the \\ 
 United States after the American War of Independence, \\ later became the second President, and is known for \\
 establishing the United States Navy? \\ 
 \textbf{Answer:} John Adams \  \textbf{Model Output:} John Q. Adams \ \ \  \\ 
 \textbf{Normalizer Output:} False \\\hline
  \rule{0pt}{2.0ex}
 \textbf{Question:}  What is the largest prime number less than or \\ equal to 100? \\ 
 \textbf{Answer:} 97\ \ \textbf{Model Output:} 98 \ \ \  \\ 
 \textbf{Normalizer Output:} False \\\bhline
\end{tabular}
\caption{Example outputs of the spelling variation normalizer. The actual examples are in Japanese, but their English translations are provided for legibility.}
\label{tab:yure_example}
\end{table}

\end{document}